\documentclass{llncs}
\usepackage[utf8]{inputenc}
\usepackage{indentfirst}
\usepackage{hyperref}
\usepackage{graphicx}
\usepackage{lmodern}  % Use a scalable font (lmodern)
\usepackage{float}
\usepackage{multirow,tabularx}
\usepackage{seqsplit}
\usepackage{amsmath}
\usepackage{regexpatch}
\usepackage{ipa} % dung cho phoneme
\usepackage{tipa} % dung cho phoneme
\usepackage{multirow} % dung cho table
\usepackage{arydshln} % Gói hỗ trợ đường nét đứt trong bảng
\usepackage{booktabs}
\usepackage{xcolor}
\makeatletter
\newcommand{\printfnsymbol}[1]{%
  \textsuperscript{\@fnsymbol{#1}}%
}

\begin{document}
\title{Whisper based Cross-Lingual Phoneme Recognition between Vietnamese and English}
\titlerunning{Whisper based Cross-Lingual Phoneme Recognition}
% If the paper title is too long for the running head, you can set
% an abbreviated paper title here
%
\author{Nguyen Huu Nhat Minh\inst{1}\thanks{Equal Contribution.}\and Tran Nguyen Anh\inst{1}\printfnsymbol{1} \and
Truong Dinh Dung\inst{1}\and
Vo Van Nam\inst{1}\and
Le Pham Tuyen\inst{2}
}
\authorrunning{Nguyen Huu Nhat Minh et. al.}
% First names are abbreviated in the running head.
% If there are more than two authors, 'et al.' is used.
%
\institute{The University of Danang, Vietnam - Korea University of Information and Communication Technology \\
\and Industrial University of Ho Chi Minh City \\
\email{nhnminh@vku.udn.vn, anhtn.21it@vku.udn.vn, dungtd.21it@vku.udn.vn, namvv.21it@vku.udn.vn, tuyen\_01036033@iuh.edu.vn}}
\maketitle              % typeset the header of the contribution
\begin{abstract}
Cross-lingual phoneme recognition has emerged as a significant challenge for accurate automatic speech recognition (ASR) when mixing Vietnamese and English pronunciations. Unlike many languages, Vietnamese relies on tonal variations to distinguish word meanings, whereas English features stress patterns and non-standard pronunciations that hinder phoneme alignment between the two languages. To address this challenge, we propose a novel bilingual speech recognition approach with two primary contributions: (1) constructing a representative bilingual phoneme set that bridges the differences between Vietnamese and English phonetic systems; (2) designing an end-to-end system that leverages the PhoWhisper pre-trained encoder for deep high-level representations to improve phoneme recognition. Our extensive experiments demonstrate that the proposed approach not only improves recognition accuracy in bilingual speech recognition for Vietnamese but also provides a robust framework for addressing the complexities of tonal and stress-based phoneme recognition. 

\end{abstract}
\section{Introduction}
Nowadays, in an increasingly interconnected and multilingual world, cross-lingual phoneme recognition has become essential in speech processing to produce a new foundation for Automatic Speech Recognition (ASR) and Text-to-Speech (TTS). Phoneme recognition involves identifying the smallest units of sound in a language, which is essential for accurate transcription and natural speech synthesis. In multilingual speech systems, phoneme recognition is the core function of ASR, helps humans better understand speeches where speakers often switch between languages, and supports TTS systems by enabling natural and clear speech synthesis across multiple languages. Cross-lingual phoneme recognition is also valuable for language learning tools, providing learners with precise feedback on their pronunciation in learning languages. In addition, advances in cross-lingual phoneme recognition have significant practical implications, particularly in the development of multilingual virtual assistants, and media platforms.

Despite notable progress in recognizing monolingual phonemes, such as those for English or Vietnamese, significant challenges remain in cross-lingual scenarios. This is particularly true for Vietnamese and English, which have fundamental differences in pronunciation. Vietnamese is a tonal language in which tones affect word meanings, whereas English is more based on stress and rhythm. The challenges are even greater when Vietnamese speakers frequently switch between languages or speak with accents influenced by their native language. For example, they often localize English words to fit Vietnamese pronunciation rules, making speech recognition more challenging for ASR systems. In addition, the task of distinguishing similar-sounding phonemes in Vietnamese and effectively handling English stress patterns poses significant difficulties. Even with large multilingual models like Whisper, current systems struggle to effectively recognize code-switching speech.
To address these issues, we propose a novel methodology for cross-lingual phoneme recognition in Vietnamese and English with the following key components:
\begin{itemize}
    \item \textbf{Constructed Cross-Lingual pronunciation set}: This set is constructed to support both standard and non-standard English pronunciations by mapping English words to Vietnamese syllables, thereby improving recognition accuracy for Vietnamese-accented English while also supporting standard English pronunciation.

    \item \textbf{Attention-based Encoder-Decoder Model} \cite{bahdanau2014neural}: This model employed the PhoWhisper encoder \cite{le2024phowhisper} and designed Transformers decoder \cite{waswani2017attention} and is trained and evaluated on large-scale Vietnamese datasets to recognize phonemes. 
\end{itemize}

In the construction of this paper, apart from the Introduction, Related works, and Conclusion sections, our approaches of cross-lingual phoneme construction and proposed model are presented through the Methodology part. Moreover, the Experiments section showcases the details of the datasets for training and evaluating the performance of the proposed model, as well as presenting the experimental results of different methods.

\section{Related Works}

In cross-lingual phoneme recognition, a significant challenge is the variability in phoneme inventories between languages. Early approaches, such as those based on acoustic features, attempted to align phonemes between languages but often struggled with inherent differences in articulatory attributes \cite{1660023}. Recent works \cite{survey-g2p} have proposed the use of articulatory features and deep learning models to better generalize across languages.
% Notably, \cite{G2P-lstm} employed an LSTM blocks as the core modules in their end-to-end system. Although this approach demonstrated effectiveness for shorter sequences, it faced challenges when dealing with long-form sequences and complex dependencies. 
The introduction of wav2vec2.0 architecture \cite{G2P-lstm} has paved the way for the success of the Allophant model \cite{glocker2023allophant}, which uses articulatory attributes to enhance cross-lingual phoneme recognition, enabling the system to better adapt to the phonetic characteristics of different languages. 
 Similarly, \cite{XLSR-53-zeroshot} also leveraged the robustness of XLSR-53 \cite{conneau2020unsupervised} framework in the zero-shot learning paradigm, yielding notable improvements in multilingual phoneme recognition tasks. 
 In addition, the study \cite{G2P-byte} introduced a combination of byte representation and Transformers-based architecture \cite{waswani2017attention}, delivering exceptional performance not only in English but also in a range of Asian languages. 
XPhoneBERT \cite{nguyen2023xphonebert} also demonstrated the potential to improve phoneme recognition by optimizing phoneme representations across languages.
In Vietnamese, phoneme recognition faces unique challenges because tones are crucial for distinguishing word meanings. The work of \cite{DaNangVMD} highlighted the importance of tone recognition in Vietnamese, noting that tonal variations greatly affect how words are understood. In English, \cite{ye2022approach} explored how acoustic and linguistic features interact to significantly improve recognition accuracy.  

Despite these advancements, there are still a very limited number of studies focusing on cross-lingual phoneme recognition between Vietnamese and English. Most current research either focuses on a single language or only considers small overlaps in phoneme systems, without fully addressing the cross-lingual challenges posed by Vietnamese tones and English stress patterns. Our work aims to bridge the gap by creating a comprehensive cross-lingual pronunciation framework, featuring a detailed and representative phonemes set, and introducing a practical framework that effectively handles the unique challenges of both languages. The proposed approach serves as a foundational framework for developing more accurate and adaptable bilingual phoneme recognition systems.

\section{Methodology}
As one of the crucial components, we constructed the pronunciation structure for each single English word using representative phonemes to effectively facilitate the capture of cross-lingual phoneme variability and similarity between Vietnamese and English. This structure forms the foundation for processing sounds from both English and Vietnamese within a unified system, simplifying the handling of speech across the two languages. Furthermore, we employed an Attention-based Encoder-Decoder to recognize phonemes. 
The model leverages the robustness of a pre-trained ASR encoder to capture phonetic features and incorporates the Transformer decoder \cite{waswani2017attention} to generate phonemes in a contextually meaningful manner.

\subsection{Constructed cross-lingual pronunciation set}
Inspired by the hierarchical structure of Vietnamese syllables described in \cite{hirarchial-vn-phoneme}, we extend the existing pronunciation set for complex phonetic components of Vietnamese Speech. As detailed in \cite{huu}, we systematically deconstructed a single syllable into three main components: \textit{Initial, Rhyme,} and \textit{Tone}. The Rhyme is further subdivided into Medial, Nucleus, and Ending. Based on this structure, we designed a vocabulary comprising $53$ phoneme categories for Vietnamese. This hierarchical structure enables the system to effectively distinguish between tones and syllables.
Additionally, previous studies have highlighted that Vietnamese speakers frequently adapt English pronunciation to align with the native phonological rules. For instance, \cite{ngo-phuong-anh} demonstrated that Vietnamese speakers tend to localize English words into the Vietnamese syllable structure due to the influence of the mother tongue. Furthermore, \cite{ngo-phuong-anh} observed that Vietnamese and English share several vowels and consonants, creating both opportunities and challenges for phoneme recognition. For example, the pair ``\texttt{\underline{p}in}" and ``\texttt{\underline{p}ie}" share the same consonant [\textipa{p}], while \cite{cross-linguistic} also pointed out that the vowel [\textipa{i}] in ``\texttt{d\underline{i}}" is phonetically similar to the vowel [\textipa{i:}] in ``\texttt{s\underline{ee}}", despite the phonological differences between two words. These findings indicate the existence of overlapping and similar phonetic features between two languages.

\begin{figure}[h]
\includegraphics[width=0.8 \textwidth]{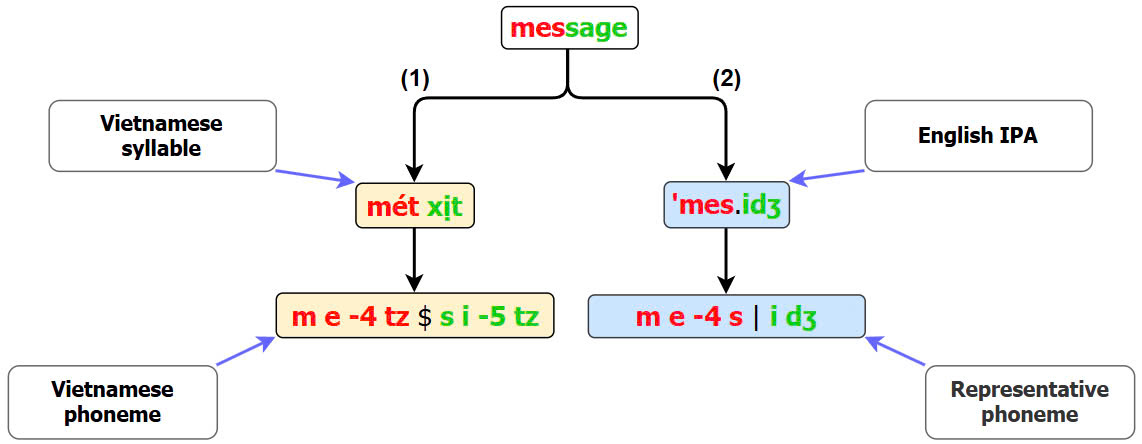}
\label{fig:pronounce_structure}
\centering
\caption{The construction of the pronunciation structure for an English word involves two processing approaches: (1) The word ``\textit{message}" is pronounced by Vietnamese speakers with non-standard pronunciation, localized to Vietnamese syllables, and (2) the word is pronounced with standard English pronunciation, transcribed as /\textit{\stress mes.id\yogh}/. The sounds from both languages are then standardized into a shared format using the proposed phoneme set. For the Vietnamese-style pronunciation, the word is broken down into ``\textit{m e -4 tz}'' and ``\textit{s i -5 tz}'', where ``\textit{-4}'' and ``\textit{-5}'' represent tone features. In contrast, the standard English pronunciation is represented as ``\textit{m e -4 s | i d\yogh}''. Furthermore, within the Vietnamese phoneme structure, ``\textit{\$}'' separates syllables, serving as spaces in the monosyllabic language, while ``\texttt{|}'' marks syllable boundaries in English.} 
\end{figure}

Building on these findings, we designed a phoneme representation vocabulary illustrated in Figure \ref{fig:pronounce_structure} to accommodate both standard and non-standard English pronunciations. For standard pronunciations, the overlapping sounds between English and Vietnamese often confuse the recognition models, leading to errors such as misclassification or hallucination when attempting to distinguish identical sounds in the two languages. To address this issue, we constructed a representative phoneme set including Vietnamese phonemes with English equivalents (e.g., the vowel /e/), and English phonemes without Vietnamese equivalents (e.g., the vowel /\ae/). This representative phoneme set ensures that English sounds are mapped to the Vietnamese phoneme equivalents, enabling the recognition model to effectively leverage shared features while reducing the confusion between the overlapped phonemes. For non-standard pronunciations, we adjusted English words to fit Vietnamese syllables by focusing on syllable adaptations. This involves representing English sounds to conform to Vietnamese's single-syllable and incorporating tones where necessary. These adjustments are beneficial to our system for handling non-standard English pronunciations, allowing the system to better process speech from users with localized accents. 

% The symbols "$" and "|" in Figure 1 essentially serve the same purpose. However, since Vietnamese is a monosyllabic language, the "$" symbol is used as a space or syllable separator. In contrast, English is a polysyllabic language, so the '|' symbol is considered a syllable separator rather than a space. Both symbols are used for presentation purposes and are not applied in the training or evaluation of the model.

% (caption... The figure [?] illustrates how this representative phoneme set works. For instance, , focusing on key sounds. This preserves important features such as Vietnamese tones and English stress, making cross-language processing for speech tasks more efficient. ...)

\subsection{Model}

% Figure of model
\begin{figure}[h]
\includegraphics[width=0.8\textwidth]{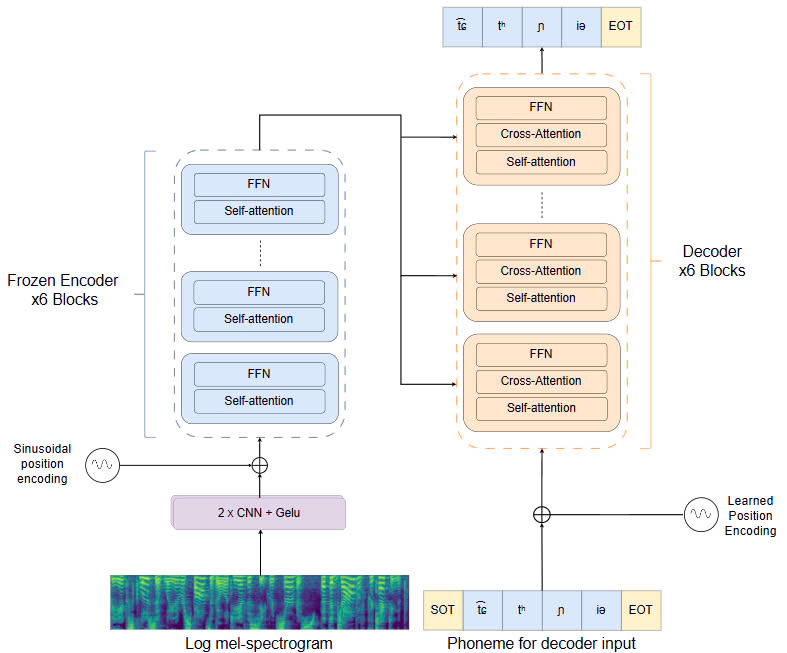}
\label{fig:bilingual_model}
\centering
\caption{The proposed framework for phoneme recognition with Vietnamese and English languages} 
\end{figure}

In this study, we utilized the Attention-Based Encoder-Decoder framework \cite{AED} for the phoneme recognition task as its superior performance in speech benchmarks. The architecture consists of two main components: PhoWhisper encoder \cite{PhoWhisper} and Transformer decoder \cite{waswani2017attention}, which work together to process audio inputs and generate phoneme-level outputs. The encoder, derived from the PhoWhisper-based model \cite{PhoWhisper}, was pre-trained on large-scale speech recognition datasets. During the training phase, its pre-trained parameters are frozen to preserve its ability to efficiently extract low-level representations from acoustic inputs. This approach ensures effective utilization of the encoder's robust learned features without additional fine-tuning. The audio input is processed into a log mel-spectrogram, resampled to $16,000$ Hz, and converted into $80$ channels magnitude representations. On the other hand, we employed the Transformer decoder \cite{waswani2017attention} and multiple cross-attention mechanisms \cite{Whisper} to preserve encoded features from the encoder. The decoder takes two inputs: an encoded representation and a sequence of previously generated tokens, enabling the model to predict the next tokens. The proposed framework is illustrated in Figure \ref{fig:bilingual_model}.

\subsubsection{Encoder}
\hfill \break
The pre-trained PhoWhisper model inherits the architecture of OpenAI's Whisper \cite{Whisper} and was fine-tuned on $843.79$ hours of Vietnamese speech dataset, encompassing a variety of accents, speech styles, and contexts. This fine-tune training allows the model to achieve remarkable proficiency in handling Vietnamese linguistic and phonetic features. Since the PhoWhisper model was trained on large-scale datasets and demonstrated outstanding performance in Vietnamese speech recognition, we leveraged its encoder to extract audio features. The input audio of the encoder is first processed into a log mel-spectrogram representation, denoted as \(X_{mel} \in \mathcal{R}^{T \times D}\), where \(T\) is the number of time frames, and \(D = 80\) is the number of mel channels. This representation is then passed through two convolutional layers (CNN), followed by the GELU activation function \cite{gelu}. Subsequently, the sinusoidal functions \(P\) provide positional information to the features after extracted by the CNN layer, enriching the features representations. These augmented features are then passed through multiple Whisper encoder blocks, each consisting of a Multi-headed-Attention (MHA) layer \cite{waswani2017attention} to capture the dependencies across the sequence and a position-wise feed-forward network (FFN), enabling the encoder to acquire contextual information at each time step.

The overall process for the encoder can be summarized as follows:
% cong thuc conv
\begin{equation}
    \begin{aligned}
        X_{\text{conv}} &= \text{GELU}(2\times\text{Conv}(X_{\text{mel}})) \\
    \end{aligned}
\end{equation}
% cong thuc PE
\begin{equation}
    X_{pos} = X_{conv} + P(X_{conv})
\end{equation}
% cong thuc MHA
\begin{equation}
    H = MHA(Q, K, V)
\end{equation}
% cong thuc MLP
\begin{equation}
    H = H + FFN(H)
\end{equation}

The encoded acoustic features are represented as a matrix \(H = [h_1, h_2, .., h_N]\), where \(N\) denotes for the length of the audio, and \(h_i\) encapsulates rich encoded acoustic features from the input audio.
\subsubsection{Decoder}
\hfill \break
While the encoder efficiently captures high-level representations of audio features, it is insufficient for generating the target sequence. Due to the complex nature of speech, the encoder alone merely represents the acoustic features and lacks the ability to generate complex phonemes based on temporal dependencies and linguistic context. To overcome this limitation, the Transformer decoder is incorporated into the architecture. The decoder operates in an autoregressive manner, where each generated token depends on both the previously generated tokens and the encoded audio features.
The input to the decoder is a sequence of phonemes represented as \(S = [s_1, s_2, .., s_N]\), where \(N\) is the length of the input sequence and \(s_i\) is an individual token. Additionally, the \texttt{<sot>} token is prepended to the start of the sequence, while the end of the sequence is marked by the \texttt{<eot>} token. The positional encoding block plays a vital role in providing positional information to individual phonemes after the embedding process. The embedded representation of the sequence \(S\) is given by:
\begin{equation}
    E = PosEnc(S) + Embedding(S)
\end{equation}
Where \(E\) is the output of the embedding layer after positional encoding is applied.

For the \textit{Self-Attention} layer, the Multi-Headed Attention module is used to capture the relationships between phonemes in a sentence. Additionally, in an end-to-end framework, while speech signals pass through a \textit{Cross-Attention} mechanism to handle the continuous and lengthy signals, facilitating the model in mapping relevant audio frames to phonetic characteristics, we observed that multiple Transformer blocks can lead to the loss of encoded information due to the  network's depth. To mitigate this, we utilized multiple cross-attention mechanisms to better preserve the encoded information throughout the model. Specifically, the \textit{Cross-Attention} mechanism is mathematically represented as:
\begin{equation}
    \textit{Cross-Attention}(Q_\text{encoder}, K, V) = softmax(\frac{Q_\text{encoder}K^{T}}{\sqrt{d_k}})V;
\end{equation}
where \(Q_{encoder}\) is the encoded audio representation, and \(K\), \(V\) are the keys and values from phoneme sequences. 

\section{Experiments}
\subsection{Datasets}
To develop and test our system for Vietnamese and English, we used a range of audio datasets representing different speakers and contexts. In this work, we compiled two types of datasets to evaluate our approach. For Vietnamese, we selected four main datasets such as: \textbf{VLSP 2020 \cite{vlsp}}, \textbf{Common Voice (CmV) \cite{cmv}}, \textbf{VIVOS \cite{vivos}}, \textbf{FOSD \cite{fosd}}.
% \begin{itemize}
%     \item \textbf{VLSP 2020 \cite{vlsp}}: is the largest dataset, accounting for $48,940$ audio samples, totaling 85.1 hours of speech. It covers various pronunciations and contexts, making it a key resource for training on Vietnamese phonetics. 
%     \item \textbf{Common Voice (CmV) \cite{cmv}}: Contributed by volunteers, this dataset contains $12,328$ samples, adding up to $13.5$ hours of speech. It includes a broad range of speakers, making it useful for capturing different voices and accents. 
%     \item \textbf{VIVOS \cite{vivos}}: Designed for Vietnamese speech tasks, this dataset consists of $8,487$ samples, totaling $10.56$ hours of speech. It provides clear examples of standard pronunciations and is valuable for training our model on common phonetic patterns. 
%     \item \textbf{FOSD \cite{fosd}}: Comprising 8,978 samples and 10.22 hours of speech, this dataset adds variety by including different speaking styles and situations.
% \end{itemize}
Together, these datasets have total 78,733 samples and 119.38 hours of speech, offering a diverse and well-rounded foundation for training our phoneme recognition system.
% \begin{table}[htb]
% \centering
% \caption{Vietnamese Data Statistics}
% \setlength{\tabcolsep}{12pt} % Điều chỉnh khoảng cách giữa các cột
% \renewcommand{\arraystretch}{1.2} % Điều chỉnh chiều cao của hàng
% \begin{tabular}{l|cc|cc}
% \hline
% \multirow{2}{*}{\textbf{Dataset}}  & \multicolumn{2}{c|}{\textbf{Training Size}}      & \multicolumn{2}{c}{\textbf{Testing Size}}        \\ \cline{2-5} 
%                           & \multicolumn{1}{c|}{\textbf{Samples}} & \textbf{Hours} & \multicolumn{1}{c|}{\textbf{Samples}} & \textbf{Hours} \\ \hline
% VIVOS                     & \multicolumn{1}{c|}{5,057}       & 6.48     & \multicolumn{1}{c|}{3,430}       & 4.08     \\ \cdashline{1-5}
% FOSD                      & \multicolumn{1}{c|}{5,499}       & 6.31     & \multicolumn{1}{c|}{3,479}       & 3.91     \\ \cdashline{1-5}
% Common Voice              & \multicolumn{1}{c|}{8,829}       & 9.68     & \multicolumn{1}{c|}{3,499}       & 3.82     \\ \cdashline{1-5}
% VLSP 2020                      & \multicolumn{1}{c|}{45,724}      & 80.73     & \multicolumn{1}{c|}{3,216}       & 4.37     \\ \hline
% \textbf{Total}                     & \multicolumn{1}{c|}{\textbf{65,109}}        & \textbf{103.2}      & \multicolumn{1}{c|}{\textbf{13,624}}        & \textbf{16.18}      \\ \hline
% \end{tabular}
% \end{table}

\begin{table}[htb]
\centering
\caption{English and Interleaved Vietnamese-English Data Statistics}
\setlength{\tabcolsep}{12pt} % Điều chỉnh khoảng cách giữa các cột
\renewcommand{\arraystretch}{1.2} % Điều chỉnh chiều cao của hàng
\begin{tabular}{l|cc|cc}
\hline
\multirow{2}{*}{\textbf{Dataset}}  & \multicolumn{2}{c|}{\textbf{\textbf{Training Size}}}             & \multicolumn{2}{c}{\textbf{Testing Size}}        \\ \cline{2-5} 
                          & \multicolumn{1}{c|}{\textbf{Samples}}     & \textbf{Hours}    & \multicolumn{1}{c|}{\textbf{Samples}} & \textbf{Hours} \\ \hline
Vietlish                  & \multicolumn{1}{c|}{3,349}       & 0.91     & \multicolumn{1}{c|}{3,137}       & 0.81     \\ \cdashline{1-5}
English Native            & \multicolumn{1}{c|}{3,349}       & 0.74     & \multicolumn{1}{c|}{3,137}       & 0.66     \\ \cdashline{1-5}
IEV            & \multicolumn{1}{c|}{5,678}       & 4.42     & \multicolumn{1}{c|}{3,110}       & 2.36     \\ \hline
\textbf{Total}                     & \multicolumn{1}{c|}{\textbf{12,376}}            & \textbf{6.07}         & \multicolumn{1}{c|}{\textbf{9,384}}        & \textbf{3,83}      \\ \hline
\end{tabular}
\end{table}

However, using only a Vietnamese speech dataset is not sufficient for our approach when applied to real-life situations, as many Vietnamese speakers tend to mix languages during conversation. They often interleave Vietnamese and English words in an interleaved manner or localize English words. To address this issue, we created a synthetic dataset consisting of native English, Vietlish, and interleaved English and Vietnamese (IEV) speech. In this dataset, the native English subset is designed for users with standard pronunciation. We selected $6,550$ commonly used English words from the Cambridge Dictionary \cite{cambridge} that frequently appear in Vietnamese conversations. The dataset is divided into three distinct subsets:
\begin{itemize}
    \item \textbf{Native English (En native)}: This subset contains English data with transcripts and audio collected directly from the Cambridge dictionary, with UK accents. 
    % The dataset consists of  3,349 samples, totaling 0.74 hours for the training set. The evaluation set includes samples different from those for training, with 3,137 samples and 0.81 hours.
    \item \textbf{Vietlish}: This subset includes only English words adapted to Vietnamese pronunciation by breaking words into syllables that align with Vietnamese phonetics. For instance, the word ``\textit{inbox}" becomes ``\texttt{in bóc}", where each English syllable is matched to a similar Vietnamese sound. 
    % The training samples are similar to those in the native English set, but the total duration is slightly higher at 0.91 hours. Additionally, the Vietlish testing set consists of 3,137 samples with a total duration of 0.66 hours.
    \item \textbf{Interleaved English and Vietnamese (IEV)}: This dataset consists of language-switching instances, where self-constructed English vocabulary is used for universal words. For instance, sentence ``\textit{Anh có mét xít cho em}." means ``I just messaged for you", the phase ``\texttt{mét xít}" corresponds to the syllables in ``\texttt{message}". 
    % The IEV dataset comprises 5,678 samples and approximately 6.07 hours for the training set. To enhance the model's robustness in handling language alternation, we evaluate its ability to recognize cross-lingual phonemes using a testing set of 3,110 samples with a total duration of 2.36 hours.
\end{itemize}
\begin{figure}[h]
\includegraphics[width=\textwidth]{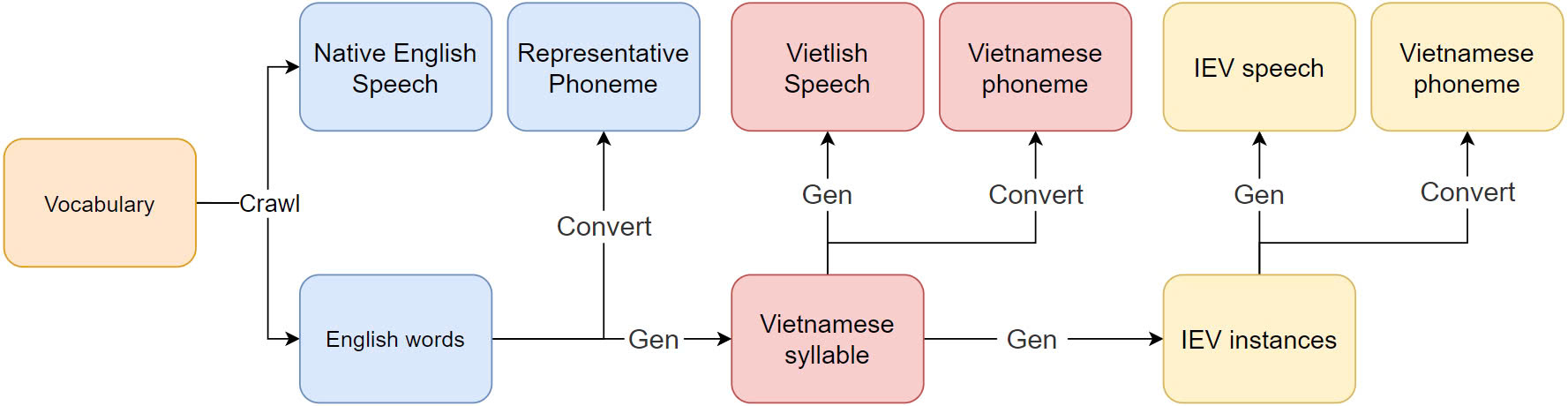}
\label{fig:flowdata}
\centering
\caption{Process of collective dataset generation.} 
\end{figure}

To construct our self-designed dataset, we first collected popular English words that could be used for the Vietlish and IEV datasets. For the native English dataset, we scraped the transcripts and corresponding audio from the Cambridge website \cite{cambridge}. Simultaneously, the English words were converted into their representative phonemes through a self-constructed linguistic conversion process. This approach helps mitigate dataset imbalance and overlap in sounds between two languages.
To create the Vietlish subset, English words were adapted into Vietnamese syllables by localizing their pronunciation to align with Vietnamese phonetics. We generated this dataset by using localized syllables and the synthetic speech service, to synthesize Vietlish speech, capturing the adapted pronunciation of English words within a Vietnamese context.
For the Interleaved English-Vietnamese (IEV) subset, English and Vietnamese words were combined into sequences, forming alternating instances. These instances were then converted into Vietnamese syllables to maintain phonetic consistency. The alternating samples were further processed using the synthetic speech service to generate speech, representing code-switching scenarios.
% commonly observed in real-life conversations. 

\subsection{Experimental Setups}
In this work, we built four models combining Wav2Vec2.0, RNN blocks, and Transformers for evaluation. While the Transformers architecture implemented for the decoder consists of six blocks, we observed that increasing the number of RNN blocks to match the number of Transformers blocks led to poor performance. 
% To address this, we limited the number of RNN blocks to three, aiming to simplify the model and prevent vanishing gradients. 
The models described as follows:
\begin{itemize}
    \item \textbf{Whisper - GRU}: integrates PhoWhisper with a Multi-Head Attention (MHA) layer followed by three GRU blocks.
    \item \textbf{Whisper - LSTM}: is similar to the Whisper-GRU architecture but replaces the GRU decoder with LSTM blocks.
    \item \textbf{Wav2Vec2.0 - Transformers}: The Wav2Vec2.0 model \cite{wav2vec} is leveraged as an encoder to capture high-level representations from audio, while six Transformers layers are implemented for the decoder.
   \item \textbf{Whisper - Transformers}: follows the architecture of an end-to-end system, where the PhoWhisper encoder is employed for extracting audio features, and a Transformers decoder, which is promising for the phoneme recognition.
\end{itemize}

The audio files were uniformly sampled at a rate of 16,000 Hz. We then derived a log mel-spectrogram using a frame shift of 20\textit{ms}, a frame length of 25\textit{ms}, and overlapping frames with a 10\textit{ms} shift. During the training phase, we employed the AdamW algorithm as an optimizer and ExponentialLR for the learning rate scheduler. We set an initial learning rate of 0.001, with a maximum of 30 epochs and a batch size of 16.

\subsection{Phoneme Error Rate}

To align the phoneme predictions, we utilized the Phoneme Error Rate (PER) methodology, which builds upon the Word Error Rate (WER) metric to calculate the ratio of incorrect predictions. We conducted our evaluation using the formula in equation \ref{equa_correct}, where ``\textit{I}'' represents insertions, “\textit{D}'' represents deletions, “\textit{S}'' represents substitutions, and “\textit{N}'' represents the total number of phonetic units:
\begin{equation} 
    \begin{aligned}
    PER &= \frac{I + D + S}{N}
    \end{aligned}
    \label{equa_correct}
\end{equation}

\begin{table}[htb]
\centering
\caption{Model Performance on Vietnamese}
\resizebox{\textwidth}{!}{%
\setlength{\tabcolsep}{8pt} % Điều chỉnh khoảng cách giữa các cột
\renewcommand{\arraystretch}{1.2} % Điều chỉnh chiều cao của hàng
\begin{tabular}{l|c|c|c|c|c}
\hline
\multirow{2}{*}{\textbf{Model (Encoder-Decoder)}} & \multirow{2}{*}{\textbf{Size}} & \multicolumn{4}{c}{\textbf{Phoneme Error Rate (\%)}} \\ \cline{3-6}
                                &                & \textbf{FOSD}   & \textbf{Vivos}   & \textbf{CmV}   & \textbf{VLSP 2020} \\ \hline
Whisper-GRU                 & 30M              & 62.72           & 46.45            & 58.79          & 77.23         \\ 
Wave2Vec-Transformer       & 152M              & 40.4            & 36.35            & 28.55          & 59.7          \\ 
Whisper-LSTM                & 32M              & 40.5            & 31.08            & 37.08          & 46.81         \\ \cdashline{1-6}
Whisper-Transformer                    & 46M              & \textbf{16.7}            & \textbf{8.85}             & \textbf{13.02}          & \textbf{22.4}          \\ \hline
\end{tabular}%
T   }
\label{table:results-vietnam}
\end{table}

\subsection{Experimental results and analysis}
\subsubsection{Vietnamese Phoneme recognition}
\hfill \break
The experimental results in \textbf{Table \ref{table:results-vietnam}} shows that the \textbf{Whisper-Transformer} significantly outperforms other models, achieving the lowest PER across all datasets with the following figures: FOSD(16.7\%), Vivos(8.85\%), CmV(13.02\%), VLSP 2020(22.4\%). This superior performance can be attributed to the well-organized architecture and its ability to efficiently capture the complex dependencies of long-form transcripts. 
Meanwhile, the opposite is true for the \textbf{Whisper-GRU} architecture, which is the smallest model with only 30M parameters. It struggles significantly, reaching the following error rates: FOSD(62.72\%), Vivos(46.45\%), CmV(58.79\%), VLSP 2020(77.23\%). This could be explained by its inability to capture complex phonetic relations and long sequences due to its simplified architecture. Additionally, the \textbf{Wav2Vec-Whisper} architecture, which has the largest parameter size (152M), shows slightly improved results, but the high error rates still exist across all datasets, with the highest one shown in VLSP 2020, where the PER is 59.7\%. This could be attributed to the model that is insufficiently tailored to Vietnamese phonetics and linguistic features, making it challenging to handle sophisticated Vietnamese acoustic features. Finally, the \textbf{Whisper-LSTM} model shows comparable performance compared to the \textbf{Wav2Vec-Transformer} on FOSD, Vivos, and CmV while the Whisper encoder-based model shows significant improvement on the VLSP dataset. 

\begin{table}[htb]
\centering
\caption{Model Performance on Synthetic Data}
\resizebox{\textwidth}{!}{%
\setlength{\tabcolsep}{8pt} % Điều chỉnh khoảng cách giữa các cột
\renewcommand{\arraystretch}{1.2} % Điều chỉnh chiều cao của hàng
\begin{tabular}{l|c|c|c|c}
\hline
\multirow{2}{*}{\textbf{Model (Encoder-Decoder)}} & \multirow{2}{*}{\textbf{Size}} & \multicolumn{3}{c}{\textbf{Phoneme Error Rate (\%)}} \\ \cline{3-5}
                                &                & \textbf{IEV}   & \textbf{Vietlish}   & \textbf{En native}   \\ \hline
Whisper-GRU                 & 30M              & 38.82           & 56.66            & 121.9           \\ 
Wave2Vec-Transformer       & 152M              & 31.79            & 33.04            & 107.7                   \\ 
Whisper-LSTM                & 32M              & 25.01            & 35.96            & 130.5                \\ \cdashline{1-5}
Whisper-Transformer                    & 46M              & \textbf{7.02}            & \textbf{16.21}             & \textbf{28.55} \\ \hline
\end{tabular}%
}
\label{table:results-mix}
\end{table}

\subsubsection{Synthetic (VN-EN) Phoneme recognition}
\hfill \break
As the results shown in \textbf{Table \ref{table:results-mix}}, which evaluate the models on the synthetic dataset, \textbf{Whisper-Transformer} again demonstrates the best performance, achieving the lowest PER across all subsets: IEV (7.02\%), Vietlish (16.21\%), and En native (28.55\%). This showcases its robustness and adaptability in handling both native English and reconstructed phonemes, as well as interleaved language patterns. On the other hand, \textbf{Whisper-GRU} performs poorly across all subsets, with PER as high as 38.82\% (IEV) and 121.9\% (En native), reflecting its limitations in modeling complex synthetic data. Similarly, \textbf{Wav2Vec-Whisper} and \textbf{Whisper-LSTM} deliver slightly better results but still fall short of addressing the intricacies of code-switching and localized phoneme structures. These findings underscore the effectiveness of \textbf{Whisper-Transformer} in generalizing across diverse linguistic patterns, both in natural and synthetic scenarios.

\section{Conclusions}

This study addresses the challenge of cross-lingual phoneme recognition between Vietnamese and English speeches. Hence, we introduced a promising methodology for constructing English word pronunciation regarding Vietnamese localization, which maps English syllables to those of Vietnamese, and representative phonemes based on the similarity of sounds. This construction allows flexibility in code-switching circumstances and localized English pronunciations commonly found in Vietnamese speech. Specifically, the proposed end-to-end architecture utilizes the robust PhoWhisper encoder, and Transformer decoder to generate phonemes efficiently. The experimental results demonstrate the effectiveness of the proposed approach, by significantly enhancing the PER results on native Vietnamese datasets compared to other architectures. The datasets used in this study primarily consist of short speech (approximately 10 seconds), as our primary goal is to focus on future short-speech conversations.

% This model illustrated superior performance across both native Vietnamese and synthetic datasets, accomplishing the lowest phoneme error rates owing to its ability to capture complex phonetic relationships and long-form sequences.

% In future work, further developments are added into our model with the aim of raising its robust applicability in real-world situations, including optimizing its learning capabilities by leveraging diverse and extensive code-switching datasets, encompassing different regional voices. 

% \renewcommand{\bibname}{References}

% \section{References}
% \begin{thebibliography}{8}
% \renewcommand{\bibname}{}
\bibliographystyle{splncs04}
\bibliography{cross-lingual.bib}
% \end{thebibliography}
% \renewcommand\bibname{References} % optional: customize title
% \bibliographystyle{splncs04}
% \begingroup
% \let\clearpage\relax  % This disables the clearpage
% \bibliography{cross-lingual}
% \endgroup

\end{document}